\documentclass[11pt, a4paper, onecolumn]{googledeepmind} %
\pdfoutput=1
\usepackage[subtle]{savetrees}

\usepackage{wrapfig}
\usepackage{listings}
\usepackage[sort&compress,numbers]{natbib}%
\usepackage{xspace}
\usepackage{setspace}
\usepackage{cleveref}
\usepackage{multirow}
\usepackage[misc,geometry]{ifsym}
\usepackage{url}

\usepackage[dvipsnames]{xcolor}

\title{Strong and Controllable 3D Motion Generation}

\author{\normalsize{Canxuan Gang}\\
AI Geeks\\
\url{https://aigeeksgroup.github.io/}}

\begin{abstract}
     \includegraphics[width=1.0\linewidth]{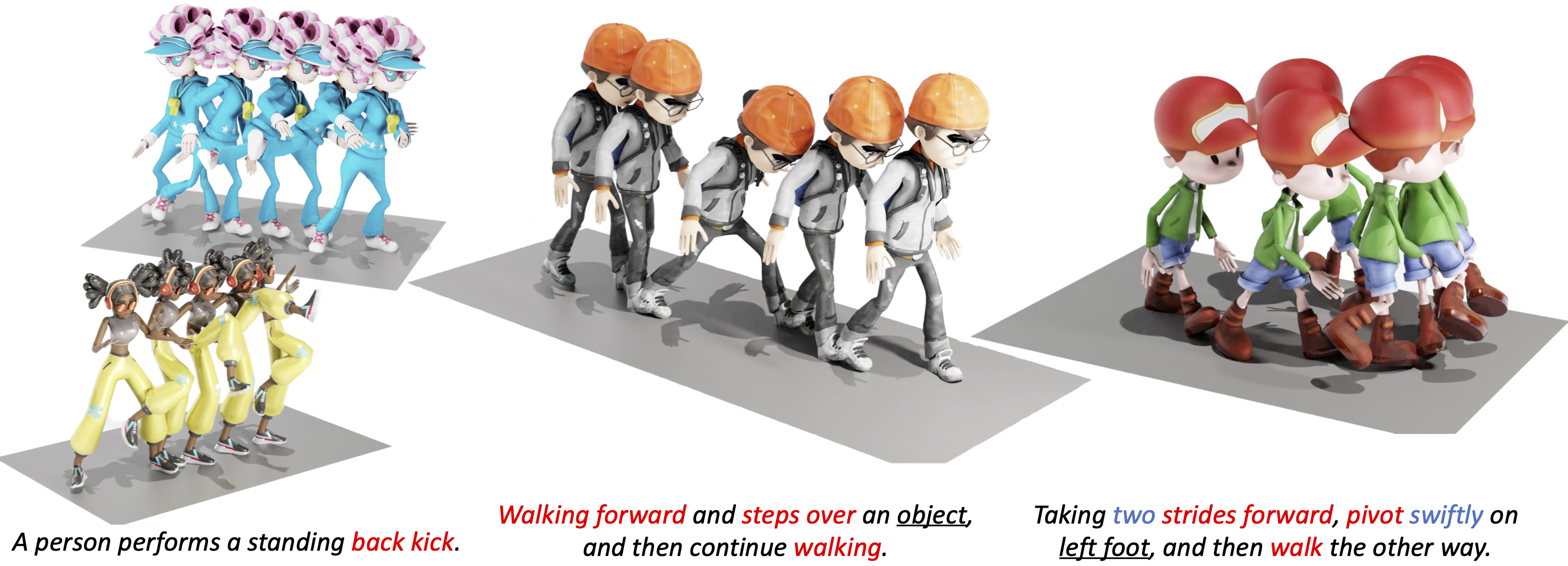}
    \captionof{figure}{The figure illustrates examples of motion generated by MoMask \cite{guo2023momask}, a state-of-the-art human motion generation model. MoMask operates using a \textit{text-to-motion} paradigm, where it takes a textual description as input and generates corresponding high-quality human motion. This approach ensures that the generated motion accurately reflects the given text conditions, showcasing MoMask's ability to produce realistic and contextually appropriate movements.}
    \label{fig:teaser}
    \vspace{2em}
 
Human motion generation is a significant pursuit in generative computer vision with widespread applications in film-making, video games, AR/VR, and human-robot interaction. Current methods mainly utilize either diffusion-based generative models or autoregressive models for text-to-motion generation. However, they face two significant challenges: (1) The generation process is time-consuming, posing a major obstacle for real-time applications such as gaming, robot manipulation, and other online settings. (2) These methods typically learn a relative motion representation guided by text, making it difficult to generate motion sequences with precise joint-level control. These challenges significantly hinder progress and limit the real-world application of human motion generation techniques. To address this gap, we propose a simple yet effective architecture consisting of two key components. Firstly, we aim to improve hardware efficiency and computational complexity in transformer-based diffusion models for human motion generation. By customizing flash linear attention, we can optimize these models specifically for generating human motion efficiently. Furthermore, we will customize the consistency model in the motion latent space to further accelerate motion generation. Secondly, we introduce Motion ControlNet, which enables more precise joint-level control of human motion compared to previous text-to-motion generation methods. These contributions represent a significant advancement for text-to-motion generation, bringing it closer to real-world applications.

\end{abstract}

\begin{document}
\maketitle

\section{Introduction}

Text-to-motion generation (T2M) \cite{guo2022generating} has garnered increasing attention due to its significant roles in various applications. Previous efforts have primarily focused on Generative Adversarial Networks (GANs) \cite{lin2018human,barsoum2018hp}, Variational Autoencoders (VAEs) \cite{ahuja2019language2pose,tevet2022motionclip,petrovich2022temos,guo2022generating,zhong2023attt2m,gong2023tm2d}, and diffusion models \cite{zhang2024motiondiffuse,tevet2022human,chen2023executing} using pairwise text-motion data, achieving impressive results. Existing approaches predominantly utilize diffusion models \cite{zhang2025fdg} as the base generative model because of their robust capability to model motion distribution. However, the current challenges in applying text-to-motion generation to real-world applications can be broadly categorized into two main folds.

Firstly, diffusion-based motion generation models \cite{zhang2024motiondiffuse,tevet2022human,chen2023executing} typically require a substantial number of sampling steps for motion synthesis during inference, even with sampling acceleration methods. For instance, MDM \cite{tevet2022human} and MLD \cite{chen2023executing} require approximately 12 seconds and 0.2 seconds, respectively, to generate a high-quality motion sequence. Thus, a significant challenge is the low efficiency that hinders the application of generating high-quality motions in various real-time scenarios.

Secondly, text-to-motion generation tasks are constrained by the need to generate motions based on text queries and conditions. Although these systems can produce high-quality and diverse motions, the generated motions are typically represented in a relative manner. This approach poses challenges for achieving precise joint-level control, which is essential for many real-world applications. The lack of precise control at the joint level limits the applicability of these systems in scenarios where exact movements and positions are crucial, such as robotics, animation, and physical therapy. Overcoming this limitation is critical for advancing the utility and effectiveness of text-to-motion generation in practical, real-world settings.

To bridge this gap, we propose a straightforward yet powerful architecture comprising two pivotal components:

\begin{itemize}
    \item Our primary focus is to enhance hardware efficiency and reduce computational complexity in transformer-based diffusion models tailored for human motion generation. We plan to achieve this by customizing flash linear attention, thus optimizing these models for efficient human motion generation. Additionally, we will tailor the consistency model within the motion latent space to further expedite the generation process.
    \item In addition to hardware optimization, we introduce Motion ControlNet, a novel framework enabling finer joint-level control over human motion compared to previous text-to-motion generation methods. This innovation marks a significant leap forward in text-to-motion generation, significantly enhancing its applicability in real-world scenarios.
\end{itemize}

\section{Related Works}

Generating human motion is a crucial task in computer vision, with applications ranging from 3D modeling to robot manipulation. A prevalent approach in this domain, known as the Text-to-Motion task, involves learning a shared latent space for both language and motion.

Several recent advancements have been made in this field. DVGAN \cite{lin2018human} utilizes a GAN \cite{goodfellow2020generative} discriminator that densely validates at each time-scale, enabling motion generation and completion. ERD-QV \cite{harvey2020robust} enhances latent representations through additive modifiers and improves transition quality with a GAN framework. HP-GAN \cite{barsoum2018hp}, trained with a modified version of WGAN-GP \cite{gulrajani2017improved}, learns a probability density function of future human poses.

Autoencoders are another notable approach, known for their robust data representation capabilities. JL2P \cite{ahuja2019language2pose} uses RNN-based autoencoders to learn a combined representation of language and pose. MotionCLIP \cite{tevet2022motionclip} reconstructs motion while aligning with corresponding text labels in the CLIP space, integrating semantic knowledge into the motion manifold. TEMOS \cite{petrovich2022temos} and T2M \cite{guo2022generating} combine Transformer-based VAE with text encoders for generating distribution parameters within the VAE latent space. AttT2M \cite{zhong2023attt2m} and TM2D \cite{gong2023tm2d} incorporate body-part spatio-temporal encoders into VQ-VAE \cite{razavi2019generating} for enhanced learning of a discrete latent space.

Diffusion models \cite{nichol2021improved, ho2020denoising, song2020denoising, dhariwal2021diffusion, rombach2022high} have recently shown promise in generating 2D images and are now being explored for motion generation. MotionDiffuse \cite{zhang2024motiondiffuse} introduces a framework for text-driven motion generation based on diffusion models, showcasing properties like probabilistic mapping and realistic synthesis. MDM \cite{tevet2022human} utilizes a classifier-free Transformer-based diffusion model for human motion prediction. MLD \cite{chen2023executing} performs diffusion in latent motion space instead of establishing connections between raw motion sequences and conditional inputs. Motion Mamba \cite{zhang2025motion}, InfiniMotion \cite{zhang2024infinimotion}, and KMM \cite{zhang2024kmm} incorporate state space models and diffusion for efficient long-sequence human motion generation. 

Transformer-based autoregressive models like MMM \cite{pinyoanuntapong2023mmm} introduce a masked motion model for generating high-quality, real-time motion with editing capabilities. MoMask \cite{guo2023momask} employs a masked transformer architecture suitable for text-driven human motion generation and similar tasks without requiring additional fine-tuning. These advancements signal promising directions in the realm of human motion generation.

\section{Objectives}

Our objective is to address two significant challenges in the field of human motion generation. First and foremost, the generation process is currently time-consuming, which presents a major obstacle for real-time applications such as gaming, robot manipulation, and various other online settings. In these environments, immediate feedback and responsiveness are crucial. The latency inherent in existing methods makes them impractical for scenarios that demand swift and seamless interactions. For instance, in gaming, slow motion generation can disrupt the player experience, while in robotic manipulation, delays can lead to inefficiencies or even failures in performing tasks. Therefore, reducing the time required for motion generation is essential to make these techniques viable for real-world applications where speed and efficiency are paramount.

Secondly, existing human motion generation methods typically rely on learning a relative motion representation guided by textual descriptions. While this approach has its advantages, it also introduces a significant limitation: the difficulty in achieving precise joint-level control in the generated motion sequences. This lack of fine-grained control can result in motions that are not accurate enough for applications that require high precision, such as medical simulations or advanced robotics. Moreover, the imprecision can hinder the creation of realistic and believable animations in virtual reality and film production. By focusing on these two challenges—reducing the generation time and enhancing joint-level control—we aim to significantly advance the field of human motion generation. Our goal is to develop techniques that not only operate efficiently in real-time but also produce highly accurate and controllable motion sequences. This will expand the potential applications of these technologies, making them more adaptable and useful in a variety of practical, real-world scenarios.

\section{Proposed Method}

\subsection{Efficient Motion Transformer}

\begin{figure}
    \centering
    \includegraphics[width=0.5\linewidth]{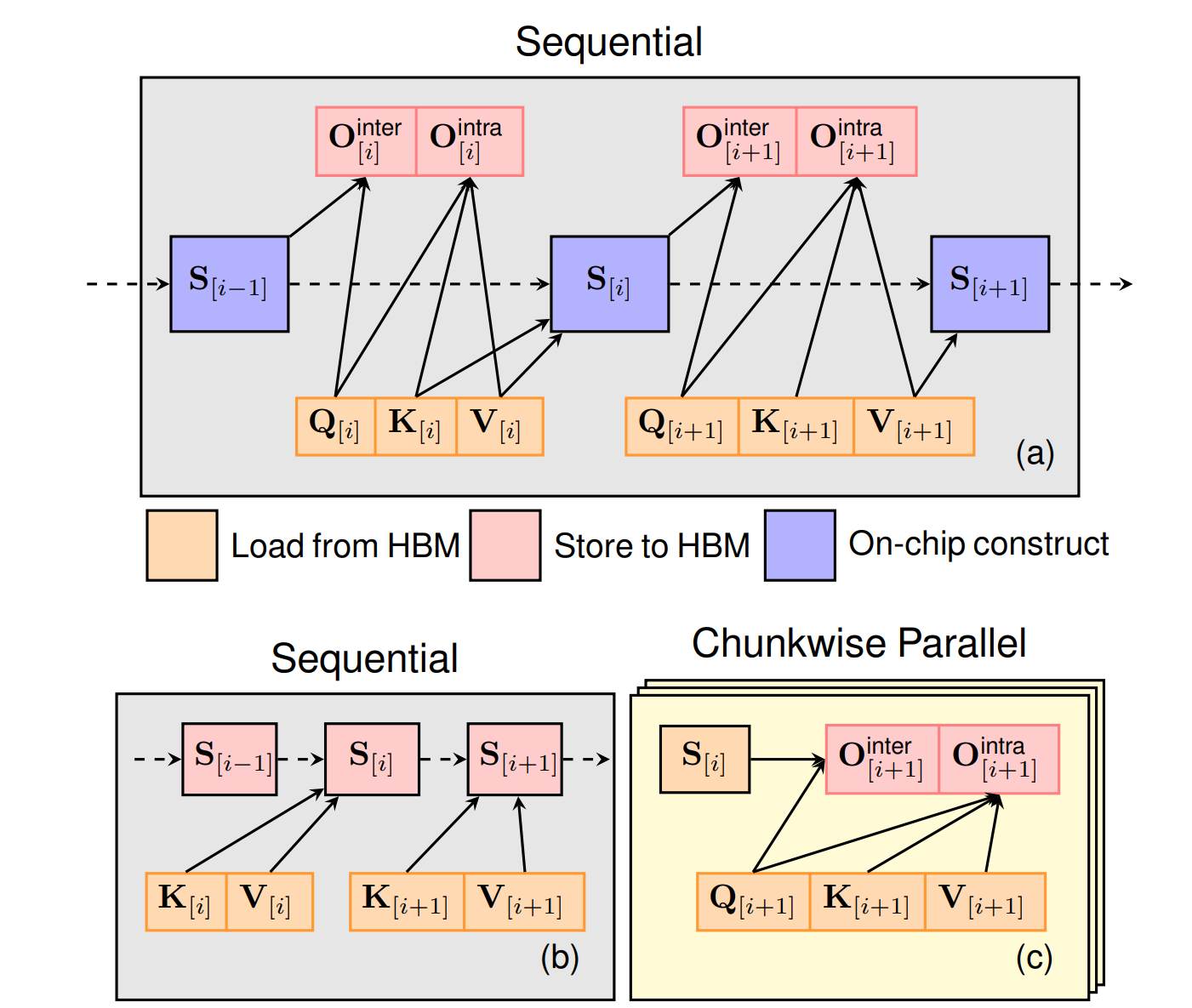}
    \caption{Comparison of Attention Mechanisms. (a) Traditional Flash Attention with quadratic complexity. (b) and (c) Flash Linear Attention, showcasing its hardware-aware design and linear complexity due to the gated attention mechanism.}
    \label{fig:fla}
\end{figure}

Efficient Motion Transformer is the foundational transformer block within our latent motion diffusion U-Net Denoiser. We have customized the promising flash linear attention to design an efficient motion transformer in the latent motion space. The \cref{fig:fla} illustrates a comparison between flash linear attention \cite{yang2023gated} and traditional flash attention \cite{dao2022flashattention, dao2023flashattention}. Flash linear attention offers several advantages, including a hardware-aware design that optimizes performance on modern computing architectures. Additionally, it features linear complexity due to its gated attention mechanism, in contrast to the quadratic complexity of traditional flash attention. This makes flash linear attention not only faster but also more scalable, enabling more efficient processing of motion data and significantly improving the overall performance of our motion generation system.

\subsection{Motion ControlNet and Latent Consistency Model}

\begin{figure}
    \centering
    \includegraphics[width=0.8\linewidth]{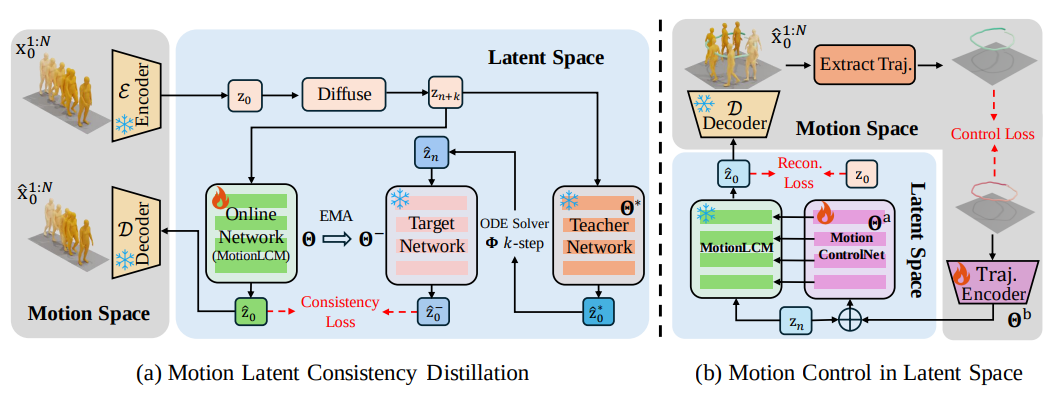}
    \caption{(a) Motion Latent Consistency Model : A consistency model customized for motion latent space, facilitating accurate prediction of clean latents through noise diffusion and network consistency. (b) Motion ControlNet: Integrated within the motion latent space, it enables controllable motion generation by supervising spatial control signals using decoded motion information.}
    \label{fig:lcm}
\end{figure}

We aim to customize Motion ControlNet within the motion latent space for more precise, even joint-level, motion control. The \cref{fig:lcm} introduces Motion ControlNet, demonstrating how it utilizes the trajectory of joints provided by users to control motion generation. Specifically, each layer in Motion ControlNet is appended with a zero-initialized linear layer to eliminate random noise during the initial training steps. This approach enhances the accuracy and control of the generated motions, ensuring they align closely with user-defined trajectories.

Motion latent consistency distillation, contributes to the acceleration of motion generation. Initially, a raw motion sequence is compressed into the latent space using a pre-trained VAE encoder. Subsequently, forward diffusion introduces noise to the latent representation. Both an online network and a teacher network then process the noisy latent representation to predict the clean latent. To ensure consistency, a loss function enforces agreement between the outputs of these networks, resulting in accurate predictions. Motion control in latent space is facilitated by integrating a Motion ControlNet, leading to controllable motion generation. Through the supervision of spatial control signals using decoded motion, this approach enhances the efficiency and precision of motion generation.

\section{Detailed Plans}

\subsection{Literature Review}

A good paper begins with a thorough literature review. This comprehensive review allows the author to understand the development trends, overarching challenges, and technical details within the research field. Given the highly competitive nature of 3D generation today, with new technologies emerging every few days and significant breakthroughs occurring monthly, it is essential to conduct a detailed literature review. This ensures access to sufficient related works and resources, such as datasets and tools, providing a solid foundation for subsequent research. By understanding the current state of the field, researchers can identify gaps, avoid duplicating existing work, and build on the latest advancements to contribute meaningfully to the domain.

\subsection{Baseline and Preparation}

Obtaining a suitable and up-to-date code base is essential for our project. During next two weeks, our goal is to identify a robust code base and thoroughly understand its inner structure. We will enhance the engineering aspects of the motion model to ensure we comprehend every technical detail. This effort will establish a solid foundation for further model design and development. By doing so, we can build on existing work effectively, avoid potential pitfalls, and ensure that our advancements are well-informed and technically sound. This foundational work is crucial for the successful implementation and innovation of our motion model.

\subsection{Design and Implement Efficient Motion Transformer}

We aim to design model with linear computation capacity by customizing flash linear attention within a Transformer-based diffusion denoiser to achieve efficient, even real-time, human motion generation. This process is tricky and challenging due to the complexity of integrating flash linear attention mechanisms with existing hardware constraints, as well as ensuring that the computations are both accurate and efficient in real-time scenarios. Additionally, fine-tuning the diffusion denoiser to work seamlessly with the hardware requires extensive testing and optimization. This work is crucial for real-world implementation, as achieving real-time motion generation can significantly enhance applications in areas such as virtual reality, gaming, and interactive simulations, making them more responsive and immersive.

\subsection{Design and Implement Motion ControlNet and Latent Consistency Model}

Our objective is to develop a motion ControlNet that enables more precise joint-level control of human motion compared to previous text-to-motion generation methods. One of the challenges we anticipate is that, unlike ControlNet's pixel space, human motion has a different latent representation that requires further exploration. Additionally, we aim to customize a consistency model within the latent space, particularly focusing on latent consistency distillation. This customization is intended to further accelerate the text-to-motion generation process. By addressing these challenges, we hope to achieve significant improvements in the precision and efficiency of motion generation models.

\subsection{Comparative and Ablation Studies}

In the upcoming weeks, we plan to conduct comprehensive experiments to thoroughly evaluate the performance of our text-to-motion generation model. This will include both comparative studies and ablation studies. We will compare our model against other state-of-the-art models using the HumanML3D dataset to benchmark its performance. Additionally, we will perform ablation studies to assess the contributions of specific components within our model. Specifically, we will evaluate the impact of the Efficient Motion Transformer, motion ControlNet, and the motion latent consistency model. By systematically disabling each component and analyzing the resulting performance, we aim to understand how each element contributes to the overall effectiveness of our model. This rigorous evaluation will provide insights into the strengths and areas for improvement in our approach, guiding further refinement and optimization.

\subsection{Paper Writing}
Our goal is to actively engage in hands-on research and complete a comprehensive paper on the latest advancements in human generation technology. We will thoroughly investigate recent improvements and innovations in the field, analyzing cutting-edge techniques and methodologies. This paper will not only document our findings but also critically assess the implications and potential applications of these advancements. By the end of the project, we aim to have a detailed and insightful overview of the current state of human generation knowledge, providing a solid basis for future research and development efforts. 

In three weeks, we will conduct a comprehensive paper on our proposed efficient motion generation architecture. This paper will detail the methodologies we employed, the challenges we encountered, and the solutions we devised. We will provide an in-depth analysis of the architecture's performance, including quantitative metrics and qualitative assessments. Additionally, the paper will include a comparison with existing models to highlight the improvements and innovations introduced by our approach. We will also discuss potential applications, future research directions, and any limitations of our work. This comprehensive documentation will serve as a valuable resource for both our team and the broader research community, ensuring transparency and facilitating further advancements in the field of motion generation. We aim to submit this academic paper in top conference including ACM MM, NeurIPS, and so on.

\subsection{Revision of Paper}

Revising the paper is a crucial step before submission. To ensure the highest quality, we will invite our senior co-author to review the paper and suggest modifications. Their expertise will help identify any weaknesses or areas for improvement, ensuring that the final version is polished and comprehensive. This collaborative revision process not only enhances the clarity and accuracy of our paper but also ensures that it meets the highest academic standards. By incorporating their feedback, we can refine our arguments, strengthen our conclusions, and present a well-rounded and thoroughly vetted paper.

\section{Acknowledgements}
I would like to acknowledge the support of Zeyu Zhang and AI Geeks in contributing to this research project.



\begin{thebibliography}{29}
\providecommand{\natexlab}[1]{#1}
\providecommand{\url}[1]{\texttt{#1}}
\expandafter\ifx\csname urlstyle\endcsname\relax
  \providecommand{\doi}[1]{doi: #1}\else
  \providecommand{\doi}{doi: \begingroup \urlstyle{rm}\Url}\fi

\bibitem[Ahuja and Morency(2019)]{ahuja2019language2pose}
Chaitanya Ahuja and Louis-Philippe Morency.
\newblock Language2pose: Natural language grounded pose forecasting.
\newblock In \emph{2019 International Conference on 3D Vision (3DV)}, pages 719--728. IEEE, 2019.

\bibitem[Barsoum et~al.(2018)Barsoum, Kender, and Liu]{barsoum2018hp}
Emad Barsoum, John Kender, and Zicheng Liu.
\newblock Hp-gan: Probabilistic 3d human motion prediction via gan.
\newblock In \emph{Proceedings of the IEEE conference on computer vision and pattern recognition workshops}, pages 1418--1427, 2018.

\bibitem[Chen et~al.(2023)Chen, Jiang, Liu, Huang, Fu, Chen, and Yu]{chen2023executing}
Xin Chen, Biao Jiang, Wen Liu, Zilong Huang, Bin Fu, Tao Chen, and Gang Yu.
\newblock Executing your commands via motion diffusion in latent space.
\newblock In \emph{Proceedings of the IEEE/CVF Conference on Computer Vision and Pattern Recognition}, pages 18000--18010, 2023.

\bibitem[Dao(2023)]{dao2023flashattention}
Tri Dao.
\newblock Flashattention-2: Faster attention with better parallelism and work partitioning.
\newblock \emph{arXiv preprint arXiv:2307.08691}, 2023.

\bibitem[Dao et~al.(2022)Dao, Fu, Ermon, Rudra, and R{\'e}]{dao2022flashattention}
Tri Dao, Dan Fu, Stefano Ermon, Atri Rudra, and Christopher R{\'e}.
\newblock Flashattention: Fast and memory-efficient exact attention with io-awareness.
\newblock \emph{Advances in Neural Information Processing Systems}, 35:\penalty0 16344--16359, 2022.

\bibitem[Dhariwal and Nichol(2021)]{dhariwal2021diffusion}
Prafulla Dhariwal and Alexander Nichol.
\newblock Diffusion models beat gans on image synthesis.
\newblock \emph{Advances in neural information processing systems}, 34:\penalty0 8780--8794, 2021.

\bibitem[Gong et~al.(2023)Gong, Lian, Chang, Guo, Jiang, Zuo, Mi, and Wang]{gong2023tm2d}
Kehong Gong, Dongze Lian, Heng Chang, Chuan Guo, Zihang Jiang, Xinxin Zuo, Michael~Bi Mi, and Xinchao Wang.
\newblock Tm2d: Bimodality driven 3d dance generation via music-text integration.
\newblock In \emph{Proceedings of the IEEE/CVF International Conference on Computer Vision}, pages 9942--9952, 2023.

\bibitem[Goodfellow et~al.(2020)Goodfellow, Pouget-Abadie, Mirza, Xu, Warde-Farley, Ozair, Courville, and Bengio]{goodfellow2020generative}
Ian Goodfellow, Jean Pouget-Abadie, Mehdi Mirza, Bing Xu, David Warde-Farley, Sherjil Ozair, Aaron Courville, and Yoshua Bengio.
\newblock Generative adversarial networks.
\newblock \emph{Communications of the ACM}, 63\penalty0 (11):\penalty0 139--144, 2020.

\bibitem[Gulrajani et~al.(2017)Gulrajani, Ahmed, Arjovsky, Dumoulin, and Courville]{gulrajani2017improved}
Ishaan Gulrajani, Faruk Ahmed, Martin Arjovsky, Vincent Dumoulin, and Aaron~C Courville.
\newblock Improved training of wasserstein gans.
\newblock \emph{Advances in neural information processing systems}, 30, 2017.

\bibitem[Guo et~al.(2022)Guo, Zou, Zuo, Wang, Ji, Li, and Cheng]{guo2022generating}
Chuan Guo, Shihao Zou, Xinxin Zuo, Sen Wang, Wei Ji, Xingyu Li, and Li~Cheng.
\newblock Generating diverse and natural 3d human motions from text.
\newblock In \emph{Proceedings of the IEEE/CVF Conference on Computer Vision and Pattern Recognition}, pages 5152--5161, 2022.

\bibitem[Guo et~al.(2023)Guo, Mu, Javed, Wang, and Cheng]{guo2023momask}
Chuan Guo, Yuxuan Mu, Muhammad~Gohar Javed, Sen Wang, and Li~Cheng.
\newblock Momask: Generative masked modeling of 3d human motions.
\newblock \emph{arXiv preprint arXiv:2312.00063}, 2023.

\bibitem[Harvey et~al.(2020)Harvey, Yurick, Nowrouzezahrai, and Pal]{harvey2020robust}
F{\'e}lix~G Harvey, Mike Yurick, Derek Nowrouzezahrai, and Christopher Pal.
\newblock Robust motion in-betweening.
\newblock \emph{ACM Transactions on Graphics (TOG)}, 39\penalty0 (4):\penalty0 60--1, 2020.

\bibitem[Ho et~al.(2020)Ho, Jain, and Abbeel]{ho2020denoising}
Jonathan Ho, Ajay Jain, and Pieter Abbeel.
\newblock Denoising diffusion probabilistic models.
\newblock \emph{Advances in neural information processing systems}, 33:\penalty0 6840--6851, 2020.

\bibitem[Lin and Amer(2018)]{lin2018human}
Xiao Lin and Mohamed~R Amer.
\newblock Human motion modeling using dvgans.
\newblock \emph{arXiv preprint arXiv:1804.10652}, 2018.

\bibitem[Nichol and Dhariwal(2021)]{nichol2021improved}
Alexander~Quinn Nichol and Prafulla Dhariwal.
\newblock Improved denoising diffusion probabilistic models.
\newblock In \emph{International conference on machine learning}, pages 8162--8171. PMLR, 2021.

\bibitem[Petrovich et~al.(2022)Petrovich, Black, and Varol]{petrovich2022temos}
Mathis Petrovich, Michael~J Black, and G{\"u}l Varol.
\newblock Temos: Generating diverse human motions from textual descriptions.
\newblock In \emph{European Conference on Computer Vision}, pages 480--497. Springer, 2022.

\bibitem[Pinyoanuntapong et~al.(2023)Pinyoanuntapong, Wang, Lee, and Chen]{pinyoanuntapong2023mmm}
Ekkasit Pinyoanuntapong, Pu~Wang, Minwoo Lee, and Chen Chen.
\newblock Mmm: Generative masked motion model.
\newblock \emph{arXiv preprint arXiv:2312.03596}, 2023.

\bibitem[Razavi et~al.(2019)Razavi, Van~den Oord, and Vinyals]{razavi2019generating}
Ali Razavi, Aaron Van~den Oord, and Oriol Vinyals.
\newblock Generating diverse high-fidelity images with vq-vae-2.
\newblock \emph{Advances in neural information processing systems}, 32, 2019.

\bibitem[Rombach et~al.(2022)Rombach, Blattmann, Lorenz, Esser, and Ommer]{rombach2022high}
Robin Rombach, Andreas Blattmann, Dominik Lorenz, Patrick Esser, and Bj{\"o}rn Ommer.
\newblock High-resolution image synthesis with latent diffusion models.
\newblock In \emph{Proceedings of the IEEE/CVF conference on computer vision and pattern recognition}, pages 10684--10695, 2022.

\bibitem[Song et~al.(2020)Song, Meng, and Ermon]{song2020denoising}
Jiaming Song, Chenlin Meng, and Stefano Ermon.
\newblock Denoising diffusion implicit models.
\newblock \emph{arXiv preprint arXiv:2010.02502}, 2020.

\bibitem[Tevet et~al.(2022{\natexlab{a}})Tevet, Gordon, Hertz, Bermano, and Cohen-Or]{tevet2022motionclip}
Guy Tevet, Brian Gordon, Amir Hertz, Amit~H Bermano, and Daniel Cohen-Or.
\newblock Motionclip: Exposing human motion generation to clip space.
\newblock In \emph{European Conference on Computer Vision}, pages 358--374. Springer, 2022{\natexlab{a}}.

\bibitem[Tevet et~al.(2022{\natexlab{b}})Tevet, Raab, Gordon, Shafir, Cohen-Or, and Bermano]{tevet2022human}
Guy Tevet, Sigal Raab, Brian Gordon, Yonatan Shafir, Daniel Cohen-Or, and Amit~H Bermano.
\newblock Human motion diffusion model.
\newblock \emph{arXiv preprint arXiv:2209.14916}, 2022{\natexlab{b}}.

\bibitem[Yang et~al.(2023)Yang, Wang, Shen, Panda, and Kim]{yang2023gated}
Songlin Yang, Bailin Wang, Yikang Shen, Rameswar Panda, and Yoon Kim.
\newblock Gated linear attention transformers with hardware-efficient training.
\newblock \emph{arXiv preprint arXiv:2312.06635}, 2023.

\bibitem[Zhang et~al.(2024{\natexlab{a}})Zhang, Cai, Pan, Hong, Guo, Yang, and Liu]{zhang2024motiondiffuse}
Mingyuan Zhang, Zhongang Cai, Liang Pan, Fangzhou Hong, Xinying Guo, Lei Yang, and Ziwei Liu.
\newblock Motiondiffuse: Text-driven human motion generation with diffusion model.
\newblock \emph{IEEE Transactions on Pattern Analysis and Machine Intelligence}, 2024{\natexlab{a}}.

\bibitem[Zhang et~al.(2025{\natexlab{a}})Zhang, Tian, Zhang, Liu, and Jin]{zhang2025fdg}
Ruicheng Zhang, Kanghui Tian, Zeyu Zhang, Qixiang Liu, and Zhi Jin.
\newblock Fdg-diff: Frequency-domain-guided diffusion framework for compressed hazy image restoration.
\newblock \emph{arXiv preprint arXiv:2501.12832}, 2025{\natexlab{a}}.

\bibitem[Zhang et~al.(2024{\natexlab{b}})Zhang, Gao, Liu, Chen, Chen, Wang, Li, and Tang]{zhang2024kmm}
Zeyu Zhang, Hang Gao, Akide Liu, Qi~Chen, Feng Chen, Yiran Wang, Danning Li, and Hao Tang.
\newblock Kmm: Key frame mask mamba for extended motion generation.
\newblock \emph{arXiv preprint arXiv:2411.06481}, 2024{\natexlab{b}}.

\bibitem[Zhang et~al.(2024{\natexlab{c}})Zhang, Liu, Chen, Chen, Reid, Hartley, Zhuang, and Tang]{zhang2024infinimotion}
Zeyu Zhang, Akide Liu, Qi~Chen, Feng Chen, Ian Reid, Richard Hartley, Bohan Zhuang, and Hao Tang.
\newblock Infinimotion: Mamba boosts memory in transformer for arbitrary long motion generation.
\newblock \emph{arXiv preprint arXiv:2407.10061}, 2024{\natexlab{c}}.

\bibitem[Zhang et~al.(2025{\natexlab{b}})Zhang, Liu, Reid, Hartley, Zhuang, and Tang]{zhang2025motion}
Zeyu Zhang, Akide Liu, Ian Reid, Richard Hartley, Bohan Zhuang, and Hao Tang.
\newblock Motion mamba: Efficient and long sequence motion generation.
\newblock In \emph{European Conference on Computer Vision}, pages 265--282. Springer, 2025{\natexlab{b}}.

\bibitem[Zhong et~al.(2023)Zhong, Hu, Zhang, and Xia]{zhong2023attt2m}
Chongyang Zhong, Lei Hu, Zihao Zhang, and Shihong Xia.
\newblock Attt2m: Text-driven human motion generation with multi-perspective attention mechanism.
\newblock In \emph{Proceedings of the IEEE/CVF International Conference on Computer Vision}, pages 509--519, 2023.

\end{thebibliography}

\end{document}